\documentclass[letterpaper, 10 pt, journal, twoside]{ieeetran}
\usepackage{amsmath,amsfonts}
\usepackage{array}
\usepackage[caption=false,font=normalsize,labelfont=sf,textfont=sf]{subfig}
\usepackage{textcomp}
\usepackage{stfloats}
\usepackage{url}
\usepackage{verbatim}
\usepackage{graphicx}

\usepackage{ragged2e}
\usepackage{hyperref}
\usepackage{graphics} 
\usepackage{times}
\usepackage{booktabs}      
\usepackage{nicefrac} 
\usepackage{microtype}
\usepackage{algorithm,multirow,xcolor}
\usepackage{algorithmicx}
\usepackage{algpseudocode}
\usepackage{mathtools}
\usepackage{cases}
\usepackage{color}
\usepackage{float}
\usepackage{bm}
\usepackage[sort]{cite}
\usepackage{tabularx}
\usepackage{threeparttable}
\usepackage{wrapfig}
\usepackage{amssymb}
\usepackage{soul}
\usepackage{epstopdf}
\usepackage{caption}
\usepackage{diagbox}
\usepackage{pifont}
\usepackage{listings}
\def\BibTeX{{\rm B\kern-.05em{\sc i\kern-.025em b}\kern-.08em
    T\kern-.1667em\lower.7ex\hbox{E}\kern-.125emX}}
\usepackage{balance}

\newcommand{\alignedtext}[1]{
  \begingroup
    \fontsize{8.5pt}{12pt}\selectfont
    \RaggedRight
    \ttfamily
    \spaceskip=.3em
    \hyphenpenalty=1000
    \exhyphenpenalty=1000
    #1
  \endgroup
}

\title{LAMARL: LLM-Aided Multi-Agent Reinforcement Learning for Cooperative Policy Generation}

\author{Guobin Zhu$^{1,2}$, Rui Zhou$^{1}$, Wenkang Ji$^{2}$, and Shiyu Zhao$^{2}$%
\thanks{Manuscript received: January, 26, 2025; Revised March, 28, 2025; Accepted May, 29, 2025. This paper was recommended for publication by Editor Kober, Jens upon evaluation of the Associate Editor and Reviewers' comments.
This work was supported by the STI 2030-Major Projects (No.2022ZD0208804), National Natural Science Foundation of China (No.62473017). (Corresponding author: Shiyu Zhao.)} 
\thanks{$^{1}$School of Automation Science and Electrical Engineering, Beihang University, Beijing, China.}%
\thanks{$^{2}$WINDY Lab, Department of Artificial Intelligence, Westlake University, Hangzhou, China.}%
\thanks{E-mail: \{zhugb, zhr\}@buaa.edu.cn, \{jiwenkang, zhaoshiyu\}@westlake. edu.cn.}
\thanks{Digital Object Identifier (DOI): see top of this page.}
}

\begin{document}
\bstctlcite{IEEEexample:BSTcontrol}
\maketitle

\begin{abstract}
Although Multi-Agent Reinforcement Learning (MARL) is effective for complex multi-robot tasks, it suffers from low sample efficiency and requires iterative manual reward tuning. Large Language Models (LLMs) have shown promise in single-robot settings, but their application in multi-robot systems remains largely unexplored. This paper introduces a novel LLM-Aided MARL (LAMARL) approach, which integrates MARL with LLMs, significantly enhancing sample efficiency without requiring manual design. LAMARL consists of two modules: the first module leverages LLMs to fully automate the generation of prior policy and reward functions. The second module is MARL, which uses the generated functions to guide robot policy training effectively. On a shape assembly benchmark, both simulation and real-world experiments demonstrate the unique advantages of LAMARL. Ablation studies show that the prior policy improves sample efficiency by an average of 185.9\% and enhances task completion, while structured prompts based on Chain-of-Thought (CoT) and basic APIs improve LLM output success rates by 28.5\%-67.5\%. Videos and code are available at \url{https://windylab.github.io/LAMARL/}
\end{abstract}

\begin{IEEEkeywords}
Multi-robot systems, Shape assembly, Multi-agent reinforcement learning, Large language models.
\end{IEEEkeywords}

\section{Introduction}
\IEEEPARstart{M}{ulti}-robot systems, through the coordination of multiple simple agents, can achieve a variety of complex tasks, such as collaborative transportation \cite{heuthe2024counterfactual,farivarnejad2022multirobot}, formation control \cite{oh2017bio,oh2015survey,zhao2019bearing}, and shape assembly \cite{sun2023mean}. Under distributed control, they also exhibit flexibility, scalability, and robustness \cite{heuthe2024counterfactual}. Hence, tremendous endeavors have been made to establish such systems. However, it is challenging to design an efficient multi-robot system without specialized expertise.

Traditional methods have mainly relied on control-theory-based designs \cite{sun2023mean,van2017distributed,10637665}, which involve precise modeling of the environment and robots, followed by careful design of control strategies to achieve optimal performance. However, these approaches are often task-specific, inflexible, and require specialized expertise, resulting in high labor and time costs~\cite{garaffa2021reinforcement}.

The development of multi-agent reinforcement learning (MARL) has recently opened new opportunities for multi-robot system research \cite{konda2020decentralized,brito2021go,garaffa2021reinforcement}. In MARL, each robot's control strategy is automatically optimized during training without requiring precise models of the environment or robots, relying instead on an appropriate reward mechanism, leading to widespread attention on MARL. With MARL, the challenge of designing complex control algorithms shifts to that of designing appropriate reward functions \cite{zhao2024mathematical}. However, manually designing reward functions for multi-robot systems is nontrivial, as it requires a deep understanding of the task logic and often requires iterative adjustments. Additionally, since policies are optimized through trial and error, MARL typically suffers from low sample efficiency.
\begin{figure}[t]
    \centering
    \includegraphics[width=1\linewidth]{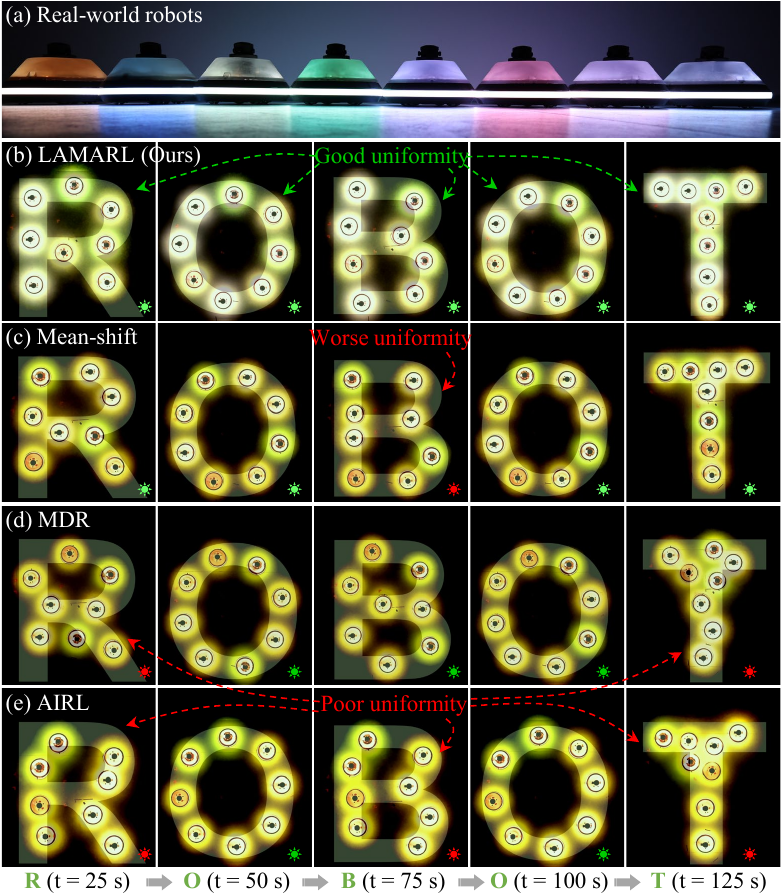}
    \caption{Real-world experimental snapshots. (a) shows eight robots arranged in a straight line. (b)-(e) illustrate the performance of four methods in assembling robots into different shapes. Green/red suns indicate satisfactory/unsatisfactory uniformity.}
    \label{fig:real_experiment}
\end{figure}

Therefore, researchers in the MARL field are increasingly focused on automating the design of rewards. Currently, approaches to this issue can be classified into three types. The first type is exploration-based methods (\emph{e.g.}, count-based, curiosity-based) \cite{yuan2023automatic,ostrovski2017count,raileanu2020ride}, which encourage agents to explore unseen states for more feedback. However, continuous exploration may hinder training convergence \cite{yuan2023automatic}. The second type is agent-based methods \cite{mareward,zheng2018learning}, which automatically learn a parametric reward by optimizing both the policy and reward simultaneously using RL algorithms, thus avoiding manual design. However, reward optimization is not a standard Markov decision process and may not guarantee the discovery of an ideal reward. The third type is expert-based methods \cite{biyik2022learning,fu2018learning,choi2011inverse}, which derive rewards from expert knowledge or data. Inverse reinforcement learning (IRL) \cite{fu2018learning,choi2011inverse} is a representative method, which can recover a reward from expert data without manual design. However, it requires a large amount of expert data to obtain an effective reward.

The rise of large language models (LLMs) has introduced a new paradigm for multi-robot system design \cite{ijcai2024p890,nayak2024longhorizon,mandi2024roco}. Leveraging their extensive knowledge, LLMs have been applied to various aspects of robotics, such as perception and decision-making \cite{ijcai2024p890}. However, most applications remain focused on single-robot scenarios, with multi-robot collaboration largely unexplored. In single-robot scenarios, LLMs are primarily applied in two ways.
The first involves deploying the LLM directly on the robot for online perception and decision-making \cite{kim2024survey}. This approach requires significant computational resources \cite{kim2024survey,chen2024scalable}, while multi-robot systems often rely on simple agents with very limited resources. Moreover, the probabilistic nature of LLMs makes the reliability and repeatability of this approach difficult to guarantee \cite{abbasi2024believe}.
The second way involves generating code offline, as demonstrated in works like Code as Policies \cite{liang2023code}, where robots execute code-based policies to perform tasks. This approach is computationally efficient and offers good reproducibility. However, it is limited to simpler tasks, and its success is not always guaranteed due to issues such as the ``hallucination'' problem with LLMs \cite{huang2023survey,zhao2023survey}.

Several recent works have integrated LLM's knowledge capabilities with RL to enhance its efficiency. For example, LLMs have been used to generate policy functions, reducing exploration in RL and assisting robots in task completion \cite{10529514}. LLMs have also been applied to generate reward functions \cite{li2024auto,ma2023eureka,xie2024textreward}, guiding robots to complete tasks without manual design. However, these approaches are limited to single-robot scenarios and typically handle only simple tasks. In contrast to single-robot systems, multi-robot systems involve agents with local perspectives and require coordination with neighbors. Given the complexity of such tasks, LLMs lack task-specific knowledge and have limited reasoning capabilities, making them prone to issues like hallucination \cite{huang2023survey,li2024challenges}.

To overcome the challenges outlined above, this paper presents a novel LLM-aided MARL approach called LAMARL that can design cooperative policies for complex multi-robot systems autonomously.
This approach can preserve MARL's ability to tackle complex tasks while leveraging LLMs' knowledge capabilities. The technical contributions are as follows.
1) We design an LLM-aided function generation module capable of outputting prior policy and reward functions. This process involves user instruction input, constraint analysis, function generation, and function review, with each step being fully automated by LLMs. It eliminates the need for manual intervention and greatly enhances design efficiency.
2) In the MARL module, the LLM-generated prior policy is incorporated into the actor loss to ensure robots possess the fundamental ability to complete tasks. Meanwhile, the LLM-generated reward is integrated into the MARL environment to guide the training of robot policies effectively.
3) To evaluate the efficiency of LAMARL, we evaluated it on the shape assembly task as shown in Fig.~\ref{fig:real_experiment}. Such a task is a long-standing, challenging benchmark in the research community~\cite{sun2023mean}. Comparative experiments in simulation and real-world experiments show that LAMARL achieves performance comparable to optimal solutions without manual design or expert data. Ablation studies further reveal that the prior policy improves sample efficiency by an average of 185.9\%, supports LLM-generated rewards in complex tasks, and that structured prompts using APIs and Chain-of-Thought (CoT) \cite{wei2022chain} enhance LLM success rates by 28.5\%-67.5\%.
To the best of our knowledge, this is the first work that combines LLMs with MARL to achieve fully autonomous policy generation for multi-robot tasks.
\begin{figure*}[!t]
    \centering
    \includegraphics[width=1\linewidth]{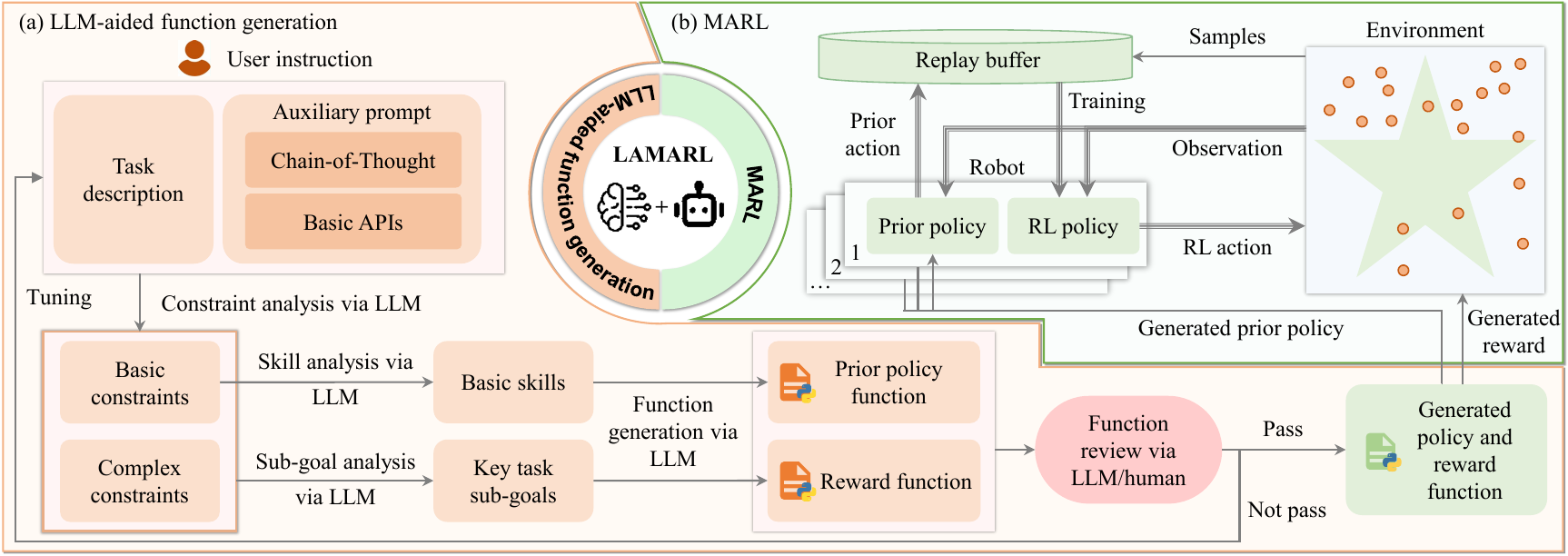}
    \caption{Method overview. (a) describes the process of using LLMs to generate the prior policy and reward functions, including user instruction input, constraint analysis, function generation, and function review. (b) describes how the prior policy and reward are integrated into MARL.}
    \label{fig:system_overview}
\end{figure*}

\section{Methodology}\label{sec:Methodology}
The method framework, as shown in Fig.~\ref{fig:system_overview}, consists of two modules. The first is the LLM-aided function generation module, which autonomously generates the prior policy and reward function using LLMs. The second is the MARL module, where the LLM-generated functions guide the training of robot policies.

\subsection{LLM-Aided Function Generation}\label{subsec:LLM-Aided Function Generation}
The LLM-aided function generation module consists of four steps. The first step is user instruction input, which consists of a task description and auxiliary prompt. Each component is expressed in natural language. The task description outlines the user's desired objectives. Auxiliary prompt provides additional context to help the LLM understand and solve the task, including CoT and basic APIs. CoT refers to a series of logically connected questions that guide the LLM to approach the task step by step. Basic APIs are pre-implemented functions that can be directly utilized. The LLM can extract deterministic information from them. Ablation experiments will show that CoT and basic APIs significantly improve function generation success rates.

The second step is constraint analysis and processing, during which the LLM addresses the questions outlined in CoT. First, the LLM analyzes the constraints that the robot may need to satisfy to complete the task. These constraints are then categorized into basic and complex constraints. Basic constraints are simple conditions with fewer steps and simpler implementation, like collision avoidance, while complex constraints involve more steps and handle intricate conditions, such as exploration. Next, the LLM identifies the basic skills the robot should possess based on the basic constraints. Additionally, the LLM identifies the key task sub-goals that must be achieved to successfully accomplish the task based on the basic and complex constraints.

The third step is function generation, including the policy and reward functions, both of which are written in Python code. At this stage, the LLM generates the policy function based on the basic skills identified in the previous step. For each basic skill, an action is created, and the output of the prior policy is the combination of all actions. This enables the policy function to execute all basic skills. Simultaneously, the LLM designs the reward function based on the identified key sub-goals. Each sub-goal corresponds to a condition, and when the condition is met, it is evaluated as true. If all conditions are true, the task is considered complete, and a reward value of 1 is returned; otherwise, the reward is 0. During this step, basic APIs are crucial for helping the LLM understand complex tasks and ensuring output accuracy.

The fourth step is function review. First, the LLM verifies whether the policy function implements the basic skills identified in Step 2. Second, the LLM checks whether the reward function evaluates all key sub-goals. If both functions pass, they are integrated into the MARL module. If not, the LLM highlights the issues and requests user adjustments. Alternatively, this step can be performed via humans.

\subsection{MARL}\label{subsec:MARL}
The MARL module consists of two parts. The first part involves integrating the LLM-generated prior policy into MARL algorithms. Using Multi-Agent Deep Deterministic Policy Gradient (MADDPG) algorithm as an example, the actions generated by the prior policy are stored in the buffer for updating the actor. Specifically, the original actor objective is to maximize $Q$, where $Q$ is the critic, but in LAMARL, it is modified to maximize $Q - \alpha (\mathbf{a} - \mathbf{a}_{\text{prior}})^2$, where $\alpha$ is the weight, $\mathbf{a}$ is the RL action and $\mathbf{a}_{\text{prior}}$ is the prior action. This regularization term ensures the RL policy mimics the prior policy and thereby equips the robots with basic task skills. Although MADDPG is used as an example, LAMARL is compatible with other policy-based algorithms as well.

The second part involves integrating the LLM-generated reward into the MARL module. This is done by incorporating the reward into the MARL environment without modifying the underlying MARL algorithm. While the prior policy equips the robots with basic task-solving abilities, it is insufficient for completing complex tasks. It is the addition of the reward that addresses this limitation. The reward function checks if all key subgoals are met, facilitating the training of a task-completing policy.
\begin{figure*}[t]
    \centering
    \includegraphics[width=1\linewidth]{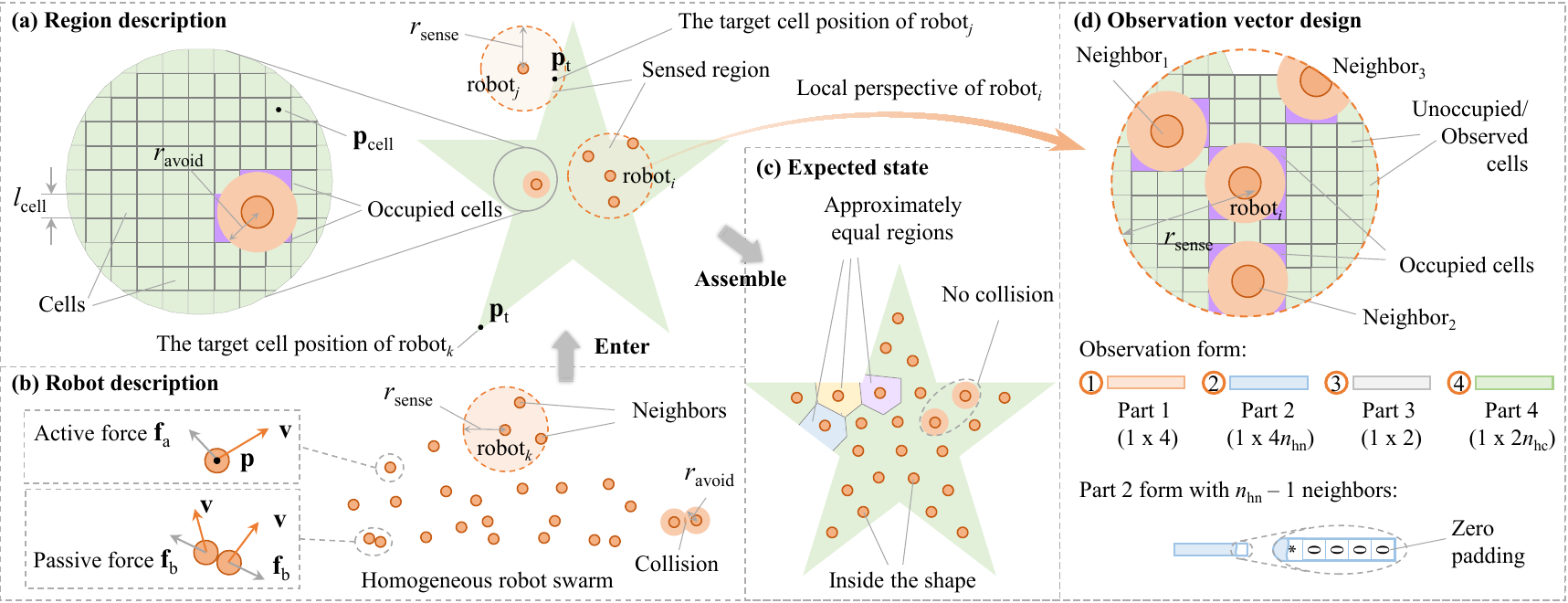}
    \caption{Environment description. (a) shows the region details, (b) shows the robot swarm setup, (c) depicts the desired assembly state, and (d) represents the observation vector design.}
    \label{fig:env_description}
\end{figure*}
\section{Task Statement}\label{sec:Task Statement}
In the following, we apply LAMARL to the shape assembly task, which also serves to demonstrate the function generation procedure. As shown in Fig.~\ref{fig:env_description}(c), this task requires a group of robots to assemble into a specified shape while maintaining equal spacing, avoiding collisions, and ensuring a uniform formation. As a longstanding benchmark in cooperative control, it presents three key challenges: first, the robots can only obtain information about the local area and neighbors; second, the robots are homogeneous, and there is no goal assignment; third, there is a global constraint that makes distributed control difficult to implement. The state-of-the-art approach achieves efficient control performance through labor-intensive design and fine-tuning \cite{sun2023mean}. However, this approach requires the total area occupied by the robots to match the target region, necessitating parameter adjustments for different region sizes and limiting adaptability.

We now address this task using LAMARL. As it involves RL, we will also describe the environment, along with the action and observation vectors design.

\subsection{Environment Description}\label{subsec:Environment Description}
\emph{\textbf{Region Description}}: The target region has a connected shape (see Fig.~\ref{fig:env_description}(a)) and is discretized into a grid composed of $n_{\text{cell}}$ cells. The side length of each cell is denoted as $l_{\text{cell}}$, and the center of each cell represents its position.

\emph{\textbf{Robot Description}}: The robot set is denoted as $\mathcal{A} = \{1, $ $\ldots, n_{\text{robot}}\}$. Each robot is represented as a disk, and its state is $\mathbf{x}$ $ = [\mathbf{p}^{T}, \mathbf{v}^{T}]^{T}$, where, $\mathbf{p}$ and $\mathbf{v}$ are the position and velocity, respectively. The robot's movement is driven by the active and passive forces (see Fig.~\ref{fig:env_description}(b)). The active force, $\mathbf{f}_{\mathrm{a}}$, is a self-generated force, which is the output of the actor network. The passive force, $\mathbf{f}_{\mathrm{b}}$, is an elastic force following Hooke's Law. Thus, the robot's dynamics is $\dot{\mathbf{p}}_i = \mathbf{v}_i, \ \dot{\mathbf{v}}_i = (\mathbf{f}_{\text{a}} + \mathbf{f}_{\text{b}})/m_i, \ i \in \mathcal{A}$, where $m_i$ is the mass of robot $i$.

Each robot has a sensing radius, $r_{\text{sense}}$, within which it can perceive neighbors and cells. When a robot perceives a neighbor, it can obtain the neighbor's state; when it perceives a cell, it can obtain the cell's position. Each robot also has a collision radius, $r_{\text{avoid}}$. A collision occurs if the inter-robot distance is less than $2r_{\text{avoid}}$, and a cell is considered occupied by a robot if the robot-to-cell distance is less than $r_{\text{avoid}}$.

\subsection{Action and Observation}\label{subsec:Action and Observation}
\emph{Action}: Each robot's action is a two-dimensional vector with components along the x and y axes. This vector indicates the active force $\mathbf{f}_{\mathrm{a}}$ exerted on the robot.

\emph{Observation}: Based on Section \ref{subsec:Environment Description}, we design the observation vector to consist of four parts (see Fig.~\ref{fig:env_description}(d)). The first part is the robot's own state, the second is the relative state of its neighbors, the third is the relative position of the target cell, and the fourth is the relative positions of unoccupied/observed cells within $r_{\text{sense}}$. The maximum number of neighbors and observed cells are denoted as $n_{\text{hn}}$ and $n_{\text{hc}}$, respectively. Hence, each robot's observation is a $(6 + 4n_{\text{hn}} + 2n_{\text{hc}})$-dimensional vector. For example, the robot $i$'s observation is $\mathbf{o}_i=[\mathbf{x}_i^T, \mathbf{x}_{ji,1}^T, \ldots, \mathbf{x}_{ji,n_{\text{hn}}}^T, \mathbf{p}_{\text{t}i}^T, \mathbf{p}_{ki,1}^T \ldots, \mathbf{p}_{ki,n_{\text{hc}}}^T]^T$, where $\mathbf{x}_{ji} = \mathbf{x}_j - \mathbf{x}_i, j \in \mathcal{N}_i$, $\mathbf{p}_{ki} = \mathbf{p}_k - \mathbf{p}_i, k \in \mathcal{C}_i$, $\mathcal{N}_i$/$\mathcal{C}_i$ is the sets of neighbors/observed cells of robot $i$. $\mathbf{p}_{\text{t}i} = \mathbf{p}_{\text{t}} - \mathbf{p}_i$, where $\mathbf{p}_{\text{t}}$ is the position of the target cell. 

Note that if the number of neighbors or observed cells within $r_{\text{sense}}$ is less than $n_{\text{hn}}$ or $n_{\text{hc}}$, respectively, the remaining elements of the observation vector are padded with zeros (see Fig.~\ref{fig:env_description}(d)). If the number of neighbors exceeds $n_{\text{hn}}$, only the $n_{\text{hn}}$ nearest neighbors are considered. If the number of observed cells exceeds $n_{\text{hc}}$, $n_{\text{hc}}$ cells are selected randomly from the set of observed cells. These adjustments ensure a fixed observation vector dimensionality.

\begin{figure*}[t]
    \centering
    \includegraphics[width=1\linewidth]{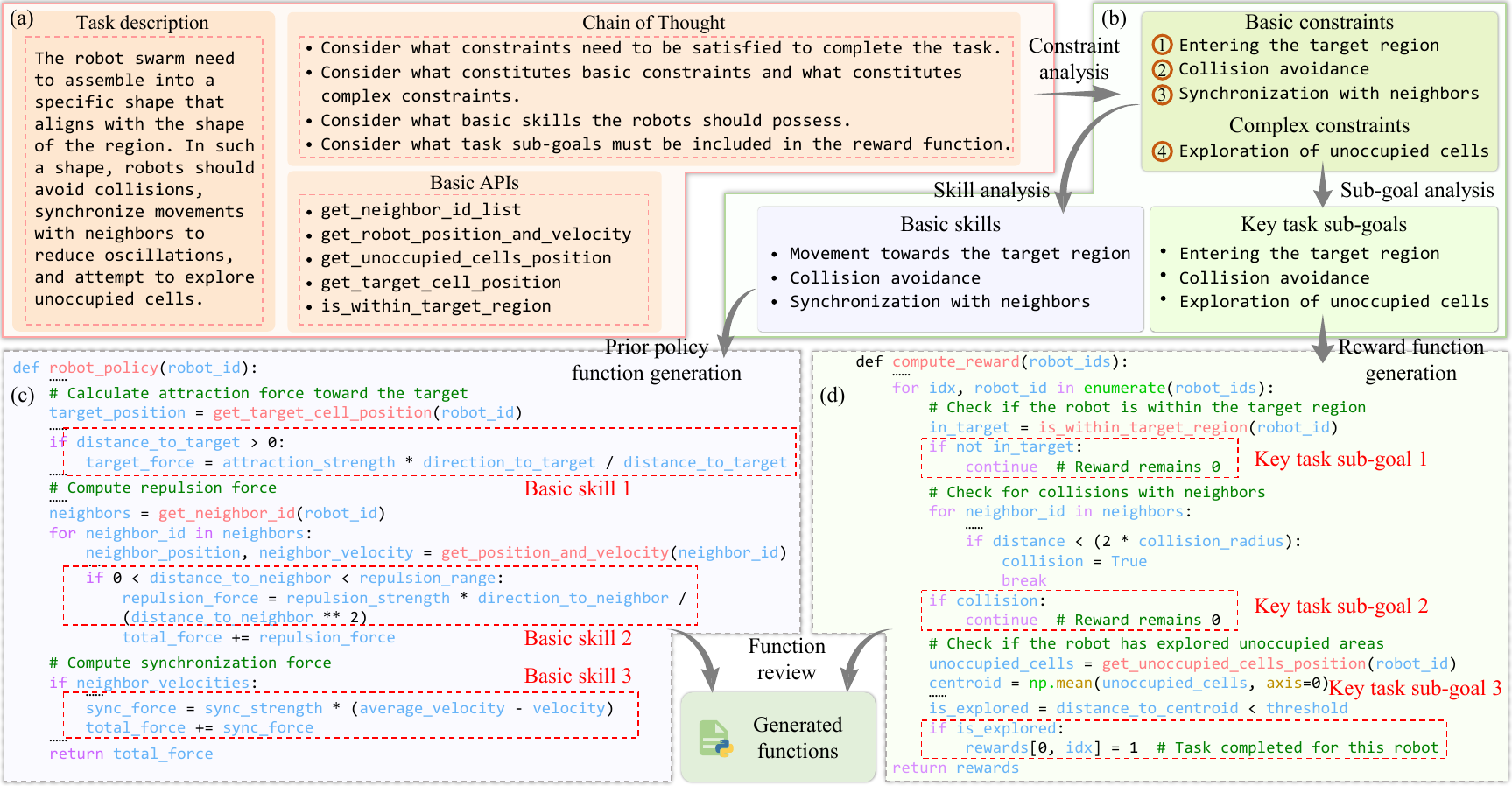}
    \caption{The workflow of the shape assembly task. (a) represents the user instruction, (b) represents constraint analysis and processing, and (c) and (d) represents the LLM-generated policy and reward function, respectively.}
    \label{fig:workflow}
\end{figure*}

\subsection{Function Generation}\label{subsec:Function Generation}
The inputs and outputs of the LLM-aided function generation module are illustrated in Fig.~\ref{fig:workflow}. The user instruction is shown in Fig.~\ref{fig:workflow}(a). Based on the user instruction, the LLM analyzes the task and determines that the following constraints may need to be satisfied to complete the task: ``\alignedtext{Entering the target region}", ``\alignedtext{Collision avoidance}", ``\alignedtext{Synchronization with neighbors}", and ``\alignedtext{Exploration of unoccupied cells}", answering the first question. It then categorizes the first three as basic constraints and the last one as a complex constraint, thus answering the second question. Next, the LLM infers that the robot's basic skills should include ``\alignedtext{Movement towards the target region}", ``\alignedtext{Collision avoidance}", and ``\alignedtext{Synchronization with \\neighbors}", answering the third question. Simultaneously, the LLM refines all constraints, identifying constraints 1, 2, and 4 as the three essential sub-goals for task completion, answering the fourth question. Under the CoT's guidance, the LLM progressively analyzes the task, and the results are intuitively correct.

Based on these basic skills and key sub-goals, the LLM designs the prior policy and reward function as shown in Fig.~\ref{fig:workflow}(c)(d). For instance, the LLM-generated prior policy includes three forces: an attraction force for entering a region, a repulsion force between neighbors, and a synchronization force for fewer oscillations. The output of the policy is the combination of these three forces. Similarly, the reward function determines whether the robot has entered the region, avoided collisions with neighbors, and completed the exploration. The task is considered complete only when all three conditions are met, at which point a reward value of 1 is returned.
Both functions passed the review and were deemed correct. Therefore, we will apply them to MARL.

\subsection{Algorithm and Training}\label{subsec:Algorithm and Training}
We adopt the MADDPG algorithm for the shape assembly task due to its support for continuous action spaces and its off-policy nature, which ensures high sample efficiency and makes it well-suited for complex tasks. MADDPG employs an actor-critic architecture \cite{lowe2017multi}. In this paper, both the actor and critic are MLPs, with Leaky-ReLU in hidden layers, Tanh for the actor's output, and no activation for the critic's output. 

Besides modifying the actor loss as described in Section \ref{subsec:MARL}, we also need to adjust the critic input. In the original MADDPG, the critic is represented as $Q(\mathbf{x}_1, \ldots, \mathbf{x}_{n_{\text{robot}}}, \mathbf{a}_1,$ $ \ldots, \mathbf{a}_{n_{\text{robot}}})$, where $\mathbf{a}_i = \mu(\mathbf{o}_i)$ and $\mu$ is the actor. The critic input includes global information. We revised it to $Q(\mathbf{o}_i, \mathbf{a}_i)$ so that the critic only receives the observation and action of the individual robot. Such a modification is necessary to support large-scale shape assembly tasks.

\section{Simulation Experiments}\label{sec:Simulation Experiments}

This section builds on the shape assembly task to examine two main aspects: a comparative experiment to highlight the advantages of LAMARL over several representative methods, and an ablation study to evaluate the importance of the prior policy and structured prompts. We begin by defining two metrics to assess each method's performance.

\subsection{Evaluation Metrics}\label{subsec:Evaluation Metrics}
\emph{Coverage rate} ($M_1$) is defined as $n_{\text{occupied}}/n_{\text{cell}}$, where $n_{\text{occupied}}$ is the total number of cells occupied by the robots. A higher $M_1$ value indicates greater area coverage by the robot swarm.

\emph{Uniformity} ($M_2$) is defined as $\sum _{i \in \mathcal{A}} (n_{\text{v},i} - n_{\text{v}})^2 / n_{\text{robot}}$, where $n_{\text{v},i}$ is the number of cells in the Voronoi region of robot $i$, and $n_{\text{v}}$ is the average value of $\{n_{\text{v},1}, \ldots, n_{\text{v},n_{\text{robot}}}\}$. A smaller $M_2$ value indicates a more uniform distribution of the robot swarm.
\begin{figure*}[t]
    \centering
    \includegraphics[width=1\linewidth]{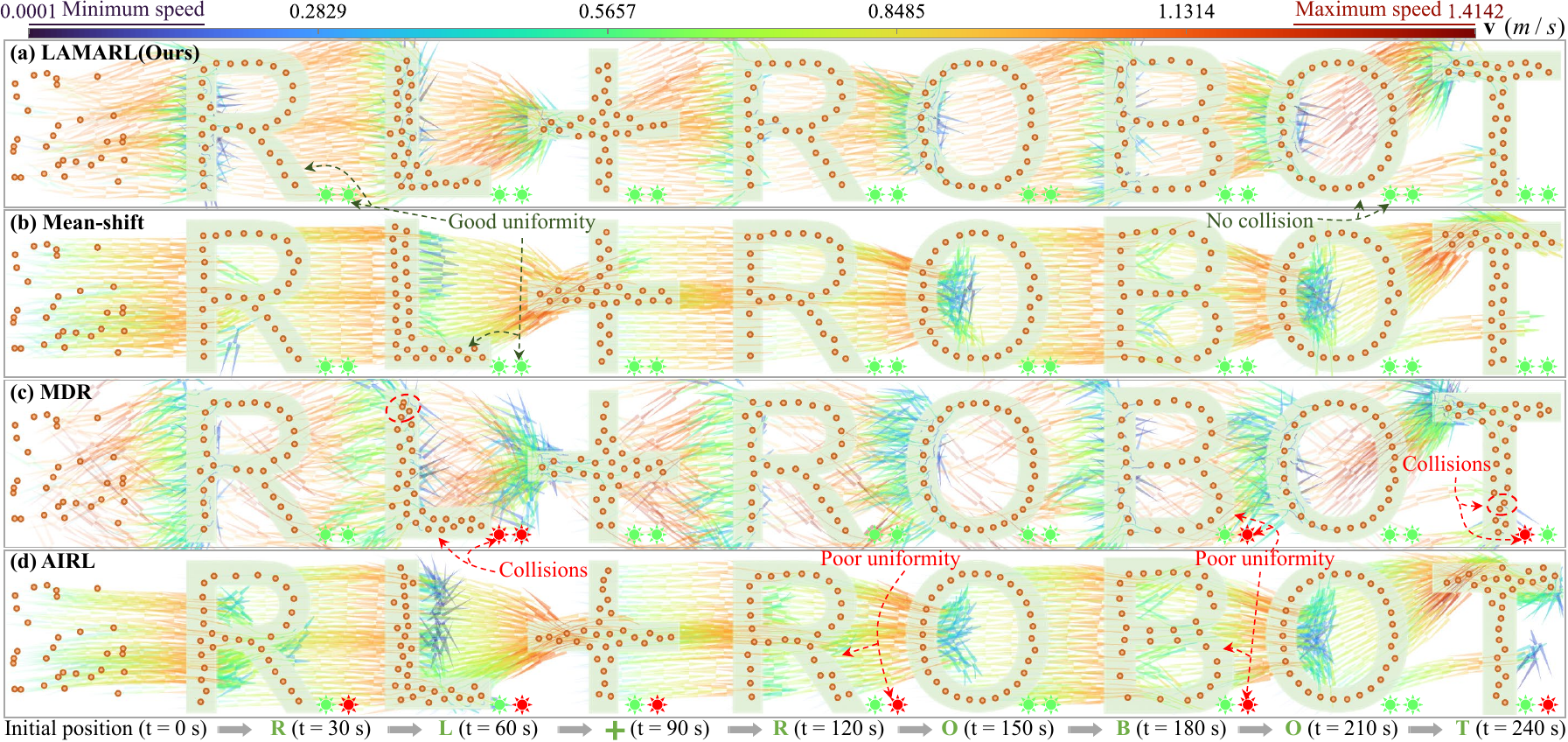}
    \caption{Comparison of shape assembly performance. A group of 30 robots start from the initial position and sequentially assemble different shapes. The curves represent the motion trajectories, with the curve colors indicating the robots' velocities. The first green/red suns indicate without/with collision, and the second green/red suns indicate satisfactory/unsatisfactory uniformity.}
    \label{fig:simulation_trajectory}
\end{figure*}

\subsection{Comparison Experiments}\label{subsec:Comparison Experiments}
\emph{\textbf{Algorithm setup}}:
We compare the LAMARL method with three representative approaches. The first approach is the traditional control method (\textbf{Mean-shift}) \cite{sun2023mean}, which includes three velocity rules: entry velocity $\mathbf{v}_{\text{ent}}$, exploration velocity $\mathbf{v}_{\text{exp}}$, and interaction velocity $\mathbf{v}_{\text{int}}$. Each rule is carefully designed by hand. The Mean-shift method currently achieves the most optimal assembly behavior, with high values of $M_1$ and $M_2$. For algorithm details, see \cite{sun2023mean}.

The second is RL with \textbf{m}anually \textbf{d}esigned \textbf{r}ewards \textbf{(MDR)}. This method uses the same action and observation forms as the LAMARL scheme, but the reward is manually specified. Inspired by the Mean-shift algorithm, the reward design considers three conditions: 1) the robot is within the shape, 2) the robot avoids collisions with neighbors, and 3) $\vert \sum _{k \in \mathcal{C}_i} \rho _k \mathbf{p}_k / \sum _{k \in \mathcal{C}_i} \rho _k - \mathbf{p}_i \vert \leq \delta, i \in \mathcal{A}$, where $\rho _k = 0.5(1+\cos \pi \frac{\Vert \mathbf{p}_k - \mathbf{p}i \Vert}{r_{\text{sense}}})$, and $\delta = 0.05$. This third condition encourages the exploration of unoccupied areas. If all conditions are met, the reward is 1; otherwise, it is 0.

The third approach is IRL, which does not require manual reward design but relies on expert data. We use the Mean-shift method as the expert policy and collect $1.5 \times 10^6$ data points of ($\mathbf{o}, \mathbf{a}$) over time to form an expert buffer. The classic \textbf{AIRL} algorithm \cite{fu2018learning} is then applied to recover the reward. AIRL consists of two loops: an inner loop for RL and an outer loop for the discriminator, which outputs the reward. While the discriminator recovers the reward, the RL loop trains the robot's policy to convergence. The discriminator's hyperparameters are set as follows: hidden dim = 180, number of hidden layers = 4, lr-discriminator = 1e-3, batch size (expert buffer) = 3072. For algorithm details, see \cite{fu2018learning}.

Note that the Mean-shift approach does not require training, whereas both MDR and AIRL involve RL training. The RL algorithms for MDR and AIRL are identical to the MADDPG used in LAMARL, except that the actor lacks a regularization term. The MADDPG hyperparameters for LAMARL, MDR, and AIRL are as follows: the number of episodes = 3000, episode length = 200, batch size = 512, hidden dim = 180, number of hidden layers = 3, lr-critic = 1e-3, lr-actor = 1e-4, exploration rate = 0.6, noise scale = 0.1, gamma = 0.99. During training, we randomize the region shape instead of fixing it, ensuring better policy adaptability.

\emph{\textbf{Task setup}}:
The setup consists of two parts. The first part involves shape parameters, where the target region shape is predefined on a canvas, and an image is generated. We can then calculate $n_{\text{cell}}$ and $l_{\text{cell}}$ based on the pixel count and side length in this image. 
The second part involves robot parameters. We set $n_{\text{robot}} = 30$, $r_{\text{sense}} = 0.4 \ \mathrm{m}$, $r_{\text{avoid}} = 0.1 \ \mathrm{m}$, $n_{\text{hn}} = 6$, and $n_{\text{hc}} = 80$. Note that the above parameters must satisfy $4n_{\text{robot}}r_{\text{avoid}}^2 \leq n_{\text{cell}}l_{\text{cell}}^2$ to ensure there is enough space within the target region to accommodate all the robots. These parameters are part of the MARL environment, while the LLM is only tasked for generating the prior policy and reward functions; therefore, they need to be manually configured.
\begin{figure*}[t]
    \centering
    \includegraphics[width=1\linewidth]{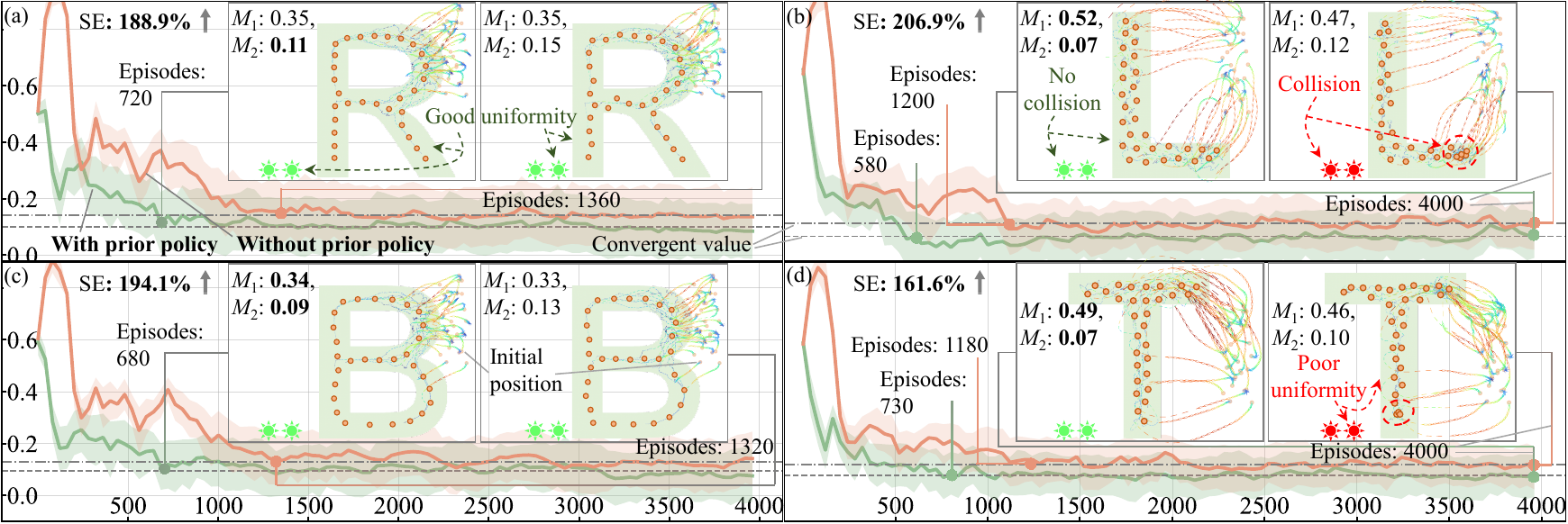}
    \caption{Ablation results of the prior policy. Each subplot's x-axis represents the number of training episodes, and the y-axis denotes uniformity $M_2$. ``SE" (sample efficiency) is calculated as the ratio of the x-coordinate at $M_2$'s first convergence with and without a prior policy. For example, for the shape ``\textbf{R}", SE = 1360 / 720 = 188.9\%.}
    \label{fig:ablation_results}
\end{figure*}

\emph{\textbf{Results comparison}}:
Fig.~\ref{fig:simulation_trajectory} illustrates the assembly results of four methods across various target shapes. Intuitively, the LAMARL method performs similarly to the Mean-shift method, with no robot collisions and a uniform distribution across the target areas. In contrast, both the MDR and AIRL methods show inadequate performance, with multiple collisions and lower uniformity. Notably, the AIRL method, influenced by expert data, performs poorly on several shapes. Table \ref{tab:adversarial_performance} summarizes the detailed $M_1$ and $M_2$ values for different shapes, where each data point represents the ``mean (standard deviation)" over 300 time steps. “Avg” indicates the average across all 26 letters; however, due to space limitations, we only present the letters corresponding to Fig. 5. It demonstrates that the LAMARL method shows comparable performance to Mean-shift, with LAMARL being less effective on certain shapes like ``\textbf{L}" and ``\textbf{T}". This is due to the relatively simple LLM-generated reward, which limits LAMARL's effectiveness in handling certain complex scenarios (\emph{e.g.}, narrow passages and multiple corners). Nevertheless, both LAMARL and Mean-shift outperform the MDR and AIRL methods overall. More importantly, the core advantage of LAMARL over the other three methods lies in the following aspects:
\begin{table}[t]
    \centering
    \caption{Detailed metrics}
    \begin{threeparttable}
    \setlength{\tabcolsep}{0.55em}{
    \renewcommand\arraystretch{1.0}
    \begin{tabular}{lllll}
        \toprule
        $M_1$ & LAMARL(Ours) & Mean-shift\cite{sun2023mean} & MDR & AIRL\cite{fu2018learning} \\
        \midrule
        “\textbf{R}” & \textbf{0.55\thinspace(0.07)} & \textbf{0.55\thinspace(0.06)} & 0.54\thinspace(0.06) & 0.53\thinspace(0.07)  \\

        “\textbf{L}” & 0.61\thinspace(0.10) & \textbf{0.63\thinspace(0.10)} & 0.55\thinspace(0.10) & 0.53\thinspace(0.12) \\

        “\ding{58}” & 0.54\thinspace(0.08) & \textbf{0.55\thinspace(0.09)} & 0.48\thinspace(0.06) & 0.49\thinspace(0.09) \\

        “\textbf{B}” & \textbf{0.54\thinspace(0.06)} & 0.53\thinspace(0.06) & 0.52\thinspace(0.07) & 0.50\thinspace(0.08) \\

        “\textbf{O}” & \textbf{0.54\thinspace(0.10)} & 0.52\thinspace(0.07) & 0.53\thinspace(0.09) & 0.53\thinspace(0.12)  \\

        “\textbf{T}” & 0.59\thinspace(0.06) & \textbf{0.60\thinspace(0.07)} & 0.57\thinspace(0.07) & 0.54\thinspace(0.07) \\

        \textbf{Avg} & \textbf{0.5837 (0.07)} & 0.5815 (0.07) & 0.5634 (0.07) & 0.5411 (0.08) \\
        
        \midrule
        $M_2$ & LAMARL(Ours) & Mean-shift\cite{sun2023mean} & MDR & AIRL\cite{fu2018learning} \\
        \midrule
        “\textbf{R}” & \textbf{0.12\thinspace(0.10)} & \textbf{0.12\thinspace(0.08)} & 0.13\thinspace(0.10) & 0.17\thinspace(0.09) \\

        “\textbf{L}” & 0.12\thinspace(0.10) & \textbf{0.10\thinspace(0.09)} & 0.15\thinspace(0.11) & 0.20\thinspace(0.13) \\

        “\ding{58}” & 0.14\thinspace(0.09) & \textbf{0.13\thinspace(0.13)} & 0.19\thinspace(0.09) & 0.20\thinspace(0.12) \\

        “\textbf{B}” & \textbf{0.12\thinspace(0.07)} & 0.13\thinspace(0.08) & 0.15\thinspace(0.11) & 0.18\thinspace(0.10) \\

        “\textbf{O}” & \textbf{0.12\thinspace(0.07)} & 0.13\thinspace(0.07) & 0.13\thinspace(0.12) & 0.13\thinspace(0.09) \\

        “\textbf{T}” & 0.10\thinspace(0.08) & \textbf{0.09\thinspace(0.08)} & 0.15\thinspace(0.08) & 0.18\thinspace(0.09) \\

        \textbf{Avg} & \textbf{0.1189 (0.08)} & 0.1211 (0.08) & 0.1533 (0.09) & 0.1821 (0.10) \\

        \bottomrule
    \end{tabular}
    }
    \end{threeparttable}
    \label{tab:adversarial_performance}
\end{table}

1) \textbf{LAMARL vs. Mean-shift method}: First, LAMARL enables fully automated policy generation without the need for manual design. Second, LAMARL offers better adaptability, as it only requires $4n_{\text{robot}}r_{\text{avoid}}^2 \leq n_{\text{cell}}l_{\text{cell}}^2$, whereas Mean-shift requires exact equality, necessitating the calculation of different $r_{\text{avoid}}$ for each shape. 2) \textbf{LAMARL vs. MDR method}: LAMARL requires no manual reward design. 3) \textbf{LAMARL vs. AIRL method}: First, LAMARL does not require expert data. Second, even when provided with large amounts of expert data, AIRL remains imitation learning by nature. Its best performance can only approach that of the Mean-shift method.

\subsection{Ablation Experiments}\label{subsec:Ablation Experiments}
This section consists of two parts: the first discusses the impact of the prior policy on algorithm performance, and the second examines the effect of a structured prompt on function generation.

\emph{\textbf{Ablation experiment 1}}:
{We conducted evaluations across all 26 letters; however, due to space limitations, we present only a subset of the results in Fig.~\ref{fig:ablation_results}. Fig.~\ref{fig:ablation_results} illustrates the variation in uniformity with and without the prior policy. For example, with the shape ``\textbf{R}", the uniformity converges after 720 episodes with the prior policy, achieving a good uniformity ($M_2 = 0.11$). In contrast, convergence is reached only after 1360 episodes without the prior policy, with a uniformity of 0.15, which is worse than 0.11. Similar results are observed for other shapes. This demonstrates that \textbf{the prior policy significantly improves sample efficiency, with an average increase of 185.9\% across all 26 letters}. Furthermore, for certain shapes like ``\textbf{L}" and ``\textbf{T}", even with 4000 episodes of training, the policy trained without the prior policy still encounters collisions and poor uniformity. This further highlights the limitations of the LLM-generated reward, while \textbf{the LLM-generated policy effectively assists in completing complex tasks}.

\emph{\textbf{Ablation experiment 2}}:
In fact, the LLM's outputs are inherently unstable, and there is no guarantee of consistently generating successful functions. On one hand, the complexity of tasks may lead to more diverse outputs. On the other hand, even with identical inputs, the LLM's outputs can vary. Hence, this ablation study examines the effect of structured prompts on the success rate of LLM-generated functions, where success is defined as the logical consistency of the generated policy and reward function with those in Fig.~\ref{fig:workflow}.

We examined the impact of removing the CoT and basic APIs components of the prompt, individually or together, over 200 trials with the OpenAI o1-preview model. The results are summarized in Table \ref{tab:ablation2}. It indicates that: 1) Without basic APIs, the success rate is very low, as the LLM struggles to effectively reason about specific and complex tasks. This highlights that \textbf{APIs can provide the LLM with fundamental, deterministic information, enhancing the LLM's accuracy}; 2) The success rate will also decrease when the LLM lacks CoT's explicit guidance on the problem-solving thought, as its reasoning ability is limited and it cannot succeed in a single step. This shows that \textbf{CoT aids task comprehension and improves accuracy}; 3) The absence of multiple components simultaneously increases the task's difficulty, further reducing the success rate.

\begin{table}
    \centering
    \caption{Ablation results of structured prompt. \textcircled{1} and \textcircled{2} indicate removing basic APIs and CoT, respectively}
    \begin{threeparttable}
    \setlength{\tabcolsep}{0.25em}{
    \renewcommand\arraystretch{1.0}
    \begin{tabular}{ccccc}
        \toprule
        Case & Full prompt & \textcircled{1} & \textcircled{2} & \textcircled{1} and \textcircled{2} \\
        \midrule
        Function generation success rate & 68.5\% & 2.50\% & 40.00\% & 1.00\% \\
        \bottomrule
    \end{tabular}
    }
    \end{threeparttable}
    \label{tab:ablation2}
\end{table}

\section{Real-world Experiments}\label{sec:Real-world Experiments}
This section aims to validate the practical feasibility of LAMARL through comparative experiments conducted on a physical platform. As shown in Fig.~\ref{fig:real_experiment}(a), we selected eight omnidirectional robots and tasked them with sequentially assembling into various shapes. All four methods were deployed on the robots for comparison. For parameter settings, only $r_{\text{avoid}}$ and $r_{\text{sense}}$ were adjusted to $r_{\text{avoid}} = 0.5 \ \mathrm{m}$ and $r_{\text{sense}} = 0.9 \ \mathrm{m}$, while all other training and task parameters remained consistent with those used in simulation.

The experimental results are presented in Fig.~\ref{fig:real_experiment}. Visually, LAMARL demonstrates superior performance, achieving well-formed and uniformly distributed shapes without collisions across all shapes. The Mean-shift method also performs well overall, except for a slight uniformity issue in shape ``\textbf{B}" likely due to the small number of robots. In contrast, the other two methods show deficiencies, particularly in assembling complex shapes such as ``\textbf{R}" and ``\textbf{T}", where the uniformity of the formations significantly deviates from the expected shapes. In summary, the conclusions from the real-world experiments align with those from the simulation experiments, thereby confirming the feasibility of LAMARL.

\section{Conclusion}\label{sec:Conclusion}
This paper proposed the LAMARL approach, which consists of two modules: the first, LLM-aided function generation, where the LLM autonomously handles the process from user input to function code; and the second, MARL, which integrates the LLM-generated functions, eliminating the need for manual design. Simulations and real-world experiments demonstrated LAMARL's unique advantages, while ablation studies highlighted the importance of prior policies and structured prompts. Indeed, the relatively simple LLM-generated reward was less effective than the carefully designed Mean-shift method for certain complex shapes, which could be the focus of future work. 

It is worth emphasizing that although this paper focuses on the shape assembly task, the proposed method demonstrates strong generalizability. This is supported by two main reasons: first, the last three steps of the LLM-aided function generation module are fully automated and task-independent; second, adapting to new tasks only requires minor changes to the CoT and APIs, since the first four CoT steps and the first two API components in Fig.~\ref{fig:workflow} are generally applicable. Extending to other tasks is also the focus of future work.

\bibliographystyle{IEEEtran}
\bibliography{IEEEabrv,main}

\begin{thebibliography}{10}
\providecommand{\url}[1]{#1}
\csname url@rmstyle\endcsname
\providecommand{\newblock}{\relax}
\providecommand{\bibinfo}[2]{#2}
\providecommand\BIBentrySTDinterwordspacing{\spaceskip=0pt\relax}
\providecommand\BIBentryALTinterwordstretchfactor{4}
\providecommand\BIBentryALTinterwordspacing{\spaceskip=\fontdimen2\font plus
\BIBentryALTinterwordstretchfactor\fontdimen3\font minus \fontdimen4\font\relax}
\providecommand\BIBforeignlanguage[2]{{%
\expandafter\ifx\csname l@#1\endcsname\relax
\typeout{** WARNING: IEEEtran.bst: No hyphenation pattern has been}%
\typeout{** loaded for the language `#1'. Using the pattern for}%
\typeout{** the default language instead.}%
\else
\language=\csname l@#1\endcsname
\fi
#2}}
\renewcommand\BIBentryALTinterwordstretchfactor{4}

\bibitem{heuthe2024counterfactual}
V.-L. Heuthe, E.~Panizon, H.~Gu, and C.~Bechinger, ``Counterfactual rewards promote collective transport using individually controlled swarm microrobots,'' \emph{Science Robotics}, vol.~9, no.~97, p. eado5888, 2024.

\bibitem{farivarnejad2022multirobot}
H.~Farivarnejad and S.~Berman, ``Multirobot control strategies for collective transport,'' \emph{Annual Review of Control, Robotics, and Autonomous Systems}, vol.~5, no.~1, pp. 205--219, 2022.

\bibitem{oh2017bio}
H.~Oh, A.~R. Shirazi, C.~Sun, and Y.~Jin, ``Bio-inspired self-organising multi-robot pattern formation: A review,'' \emph{Robotics and Autonomous Systems}, vol.~91, pp. 83--100, 2017.

\bibitem{oh2015survey}
K.-K. Oh, M.-C. Park, and H.-S. Ahn, ``A survey of multi-agent formation control,'' \emph{Automatica}, vol.~53, pp. 424--440, 2015.

\bibitem{zhao2019bearing}
S.~Zhao and D.~Zelazo, ``Bearing rigidity theory and its applications for control and estimation of network systems: Life beyond distance rigidity,'' \emph{IEEE Control Systems Magazine}, vol.~39, no.~2, pp. 66--83, 2019.

\bibitem{sun2023mean}
G.~Sun, \emph{et~al.}, ``Mean-shift exploration in shape assembly of robot swarms,'' \emph{Nature Communications}, vol.~14, no.~1, p. 3476, 2023.

\bibitem{van2017distributed}
R.~Van~Parys and G.~Pipeleers, ``Distributed mpc for multi-vehicle systems moving in formation,'' \emph{Robotics and Autonomous Systems}, vol.~97, pp. 144--152, 2017.

\bibitem{10637665}
P.~Roque, P.~Miraldo, and D.~V. Dimarogonas, ``Multi-agent formation control using epipolar constraints,'' \emph{IEEE Robotics and Automation Letters}, vol.~9, no.~12, pp. 11\,002--11\,009, 2024.

\bibitem{garaffa2021reinforcement}
L.~C. Garaffa, M.~Basso, A.~A. Konzen, and E.~P. de~Freitas, ``Reinforcement learning for mobile robotics exploration: A survey,'' \emph{IEEE Transactions on Neural Networks and Learning Systems}, vol.~34, no.~8, pp. 3796--3810, 2021.

\bibitem{konda2020decentralized}
R.~Konda, H.~M. La, and J.~Zhang, ``Decentralized function approximated q-learning in multi-robot systems for predator avoidance,'' \emph{IEEE Robotics and Automation Letters}, vol.~5, no.~4, pp. 6342--6349, 2020.

\bibitem{brito2021go}
B.~Brito, M.~Everett, J.~P. How, and J.~Alonso-Mora, ``Where to go next: Learning a subgoal recommendation policy for navigation in dynamic environments,'' \emph{IEEE Robotics and Automation Letters}, vol.~6, no.~3, pp. 4616--4623, 2021.

\bibitem{zhao2024mathematical}
S.~Zhao, \emph{Mathematical Foundations of Reinforcement Learning}.\hskip 1em plus 0.5em minus 0.4em\relax Springer Nature Press, 2024.

\bibitem{yuan2023automatic}
M.~Yuan, B.~Li, X.~Jin, and W.~Zeng, ``Automatic intrinsic reward shaping for exploration in deep reinforcement learning,'' in \emph{Proceedings of the International Conference on Machine Learning}, 2023, pp. 40\,531--40\,554.

\bibitem{ostrovski2017count}
G.~Ostrovski, M.~G. Bellemare, A.~Oord, and R.~Munos, ``Count-based exploration with neural density models,'' in \emph{Proceedings of the International Conference on Machine Learning}, 2017, pp. 2721--2730.

\bibitem{raileanu2020ride}
R.~Raileanu and T.~Rocktäschel, ``Ride: Rewarding impact-driven exploration for procedurally-generated environments,'' in \emph{Proceedings of the International Conference on Learning Representations}, 2020.

\bibitem{mareward}
H.~Ma, K.~Sima, T.~V. Vo, D.~Fu, and T.-Y. Leong, ``Reward shaping for reinforcement learning with an assistant reward agent,'' in \emph{Proceedings of the International Conference on Machine Learning}, 2024, pp. 33\,925--33\,939.

\bibitem{zheng2018learning}
Z.~Zheng, J.~Oh, and S.~Singh, ``On learning intrinsic rewards for policy gradient methods,'' in \emph{Proceedings of the Annual Conference on Neural Information Processing Systems}, vol.~31, 2018.

\bibitem{biyik2022learning}
E.~B{\i}y{\i}k, D.~P. Losey, M.~Palan, N.~C. Landolfi, G.~Shevchuk, and D.~Sadigh, ``Learning reward functions from diverse sources of human feedback: Optimally integrating demonstrations and preferences,'' \emph{The International Journal of Robotics Research}, vol.~41, no.~1, pp. 45--67, 2022.

\bibitem{fu2018learning}
J.~Fu, K.~Luo, and S.~Levine, ``Learning robust rewards with adverserial inverse reinforcement learning,'' in \emph{Proceedings of the International Conference on Learning Representations}, 2018.

\bibitem{choi2011inverse}
J.-D. Choi and K.-E. Kim, ``Inverse reinforcement learning in partially observable environments,'' \emph{Journal of Machine Learning Research}, vol.~12, pp. 691--730, 2011.

\bibitem{ijcai2024p890}
T.~Guo, \emph{et~al.}, ``Large language model based multi-agents: A survey of progress and challenges,'' in \emph{Proceedings of the International Joint Conference on Artificial Intelligence}, 2024, pp. 8048--8057.

\bibitem{nayak2024longhorizon}
S.~Nayak, \emph{et~al.}, ``Long-horizon planning for multi-agent robots in partially observable environments,'' in \emph{Multi-modal Foundation Model meets Embodied AI Workshop @ ICML2024}, 2024.

\bibitem{mandi2024roco}
Z.~Mandi, S.~Jain, and S.~Song, ``Roco: Dialectic multi-robot collaboration with large language models,'' in \emph{Proceedings of the IEEE International Conference on Robotics and Automation}, 2024, pp. 286--299.

\bibitem{kim2024survey}
Y.~Kim, D.~Kim, J.~Choi, J.~Park, N.~Oh, and D.~Park, ``A survey on integration of large language models with intelligent robots,'' \emph{Intelligent Service Robotics}, vol.~17, no.~5, pp. 1091--1107, 2024.

\bibitem{chen2024scalable}
Y.~Chen, J.~Arkin, Y.~Zhang, N.~Roy, and C.~Fan, ``Scalable multi-robot collaboration with large language models: Centralized or decentralized systems?'' in \emph{Proceedings of the IEEE International Conference on Robotics and Automation}, 2024, pp. 4311--4317.

\bibitem{abbasi2024believe}
Y.~Abbasi-Yadkori, I.~Kuzborskij, A.~Gy{\"o}rgy, and C.~Szepesvari, ``To believe or not to believe your {LLM}: Iterativeprompting for estimating epistemic uncertainty,'' in \emph{Proceedings of the Annual Conference on Neural Information Processing Systems}, 2024.

\bibitem{liang2023code}
J.~Liang, \emph{et~al.}, ``Code as policies: Language model programs for embodied control,'' in \emph{Proceedings of the IEEE International Conference on Robotics and Automation}, 2023, pp. 9493--9500.

\bibitem{huang2023survey}
L.~Huang, \emph{et~al.}, ``A survey on hallucination in large language models: Principles, taxonomy, challenges, and open questions,'' \emph{ACM Transactions on Information Systems}, 2023.

\bibitem{zhao2023survey}
W.~X. Zhao, \emph{et~al.}, ``A survey of large language models,'' \emph{arXiv preprint arXiv:2303.18223}, 2023.

\bibitem{10529514}
L.~Chen, Y.~Lei, S.~Jin, Y.~Zhang, and L.~Zhang, ``Rlingua: Improving reinforcement learning sample efficiency in robotic manipulations with large language models,'' \emph{IEEE Robotics and Automation Letters}, vol.~9, no.~7, pp. 6075--6082, 2024.

\bibitem{li2024auto}
H.~Li, \emph{et~al.}, ``Auto mc-reward: Automated dense reward design with large language models for minecraft,'' in \emph{Proceedings of the IEEE/CVF Conference on Computer Vision and Pattern Recognition}, 2024, pp. 16\,426--16\,435.

\bibitem{ma2023eureka}
Y.~J. Ma, \emph{et~al.}, ``Eureka: Human-level reward design via coding large language models,'' in \emph{Proceedings of the International Conference on Learning Representations}, 2024.

\bibitem{xie2024textreward}
T.~Xie, \emph{et~al.}, ``Text2reward: Reward shaping with language models for reinforcement learning,'' in \emph{Proceedings of the International Conference on Learning Representations}, 2024.

\bibitem{li2024challenges}
P.~Li, V.~Menon, B.~Gudiguntla, D.~Ting, and L.~Zhou, ``Challenges faced by large language models in solving multi-agent flocking,'' \emph{arXiv preprint arXiv:2404.04752}, 2024.

\bibitem{wei2022chain}
J.~Wei, \emph{et~al.}, ``Chain-of-thought prompting elicits reasoning in large language models,'' in \emph{Proceedings of the Annual Conference on Neural Information Processing Systems}, 2022, pp. 24\,824--24\,837.

\bibitem{lowe2017multi}
R.~Lowe, Y.~I. Wu, A.~Tamar, J.~Harb, O.~Pieter~Abbeel, and I.~Mordatch, ``Multi-agent actor-critic for mixed cooperative-competitive environments,'' in \emph{Proceedings of the Annual Conference on Neural Information Processing Systems}, 2017, pp. 6382--6393.

\end{thebibliography}

\end{document}